\documentclass{article}

\usepackage{arxiv}

\usepackage[utf8]{inputenc} 
\usepackage[T1]{fontenc}    
\usepackage{hyperref}       
\usepackage{url}            
\usepackage{booktabs}       
\usepackage{amsfonts}       
\usepackage{nicefrac}       
\usepackage{microtype}      
\usepackage{lipsum}

\usepackage{graphicx}
\usepackage{amsthm}
\usepackage{amsmath}
\theoremstyle{definition}
\newtheorem{defn}{Definition}
\usepackage{booktabs}
 \usepackage{enumitem}
\usepackage{amssymb}
\usepackage{bbm}
\usepackage{wrapfig}

\title{GripNet: Graph Information Propagation on Supergraph for Heterogeneous Graphs}

\author{
 Hao Xu$^1$, Shengqi Sang$^{2,3}$, Peizhen Bai$^{4}$, Laurence Yang$^1$, Haiping Lu$^{4}$  \\
$^1$Queen's University, $^2$University of Waterloo, $^3$Perimeter Institute for Theoretical Physics, $^4$University of Sheffield \\
\texttt{\{xu.hao,laurence.yang\}@queensu.ca}, \texttt{s4sang@uwaterloo.ca}, \texttt{\{pbai2,h.lu\}@sheffield.ac.uk} 
}

\begin{document}
\maketitle
\begin{abstract}
Heterogeneous graph representation learning aims to learn low-dimensional vector representations of different types of entities and relations to empower downstream tasks. Existing methods either capture semantic relationships but indirectly leverage node/edge attributes in a complex way, or leverage node/edge attributes directly without taking semantic relationships into account. When involving multiple convolution operations, they also have poor scalability. To overcome these limitations, this paper proposes a flexible and efficient \textbf{Gr}aph \textbf{i}nformation \textbf{p}ropagation \textbf{Net}work (GripNet) framework. Specifically, we introduce a new \textit{supergraph} data structure consisting of supervertices and superedges. A supervertex is a semantically-coherent subgraph. A superedge defines an information propagation path between two supervertices. GripNet learns new representations for the supervertex of interest by propagating information along the defined path using multiple layers. We construct multiple large-scale graphs and evaluate GripNet against competing methods to show its superiority in link prediction, node classification, and data integration.
\end{abstract}


\section{Introduction}
\label{sec:intro}

A heterogeneous graph/network contains multiple types of nodes and relations to represent a wide range of real-world data such as information \cite{yang2014embedding, yun2019graph}, social \cite{wu2019graph, zhang2015geosoca}, biomedical \cite{zitnik2018modeling} and chemical \cite{gilmer2017message} networks. 
Modelling its node/relation properties is important in various machine learning (ML) tasks, e.g., node classification/clustering \cite{dong2017metapath2vec, yun2019graph, sun2019vgraph}, knowledge graph completion \cite{schlichtkrull2018modeling, yang2014embedding}, link prediction \cite{zitnik2018modeling}, and recommendation \cite{wu2019graph}.
Graph representation learning (GRL), a.k.a. graph embedding, is a popular solution that embeds entities and/or relations into a low dimensional vector space with the topological information and structure of graph preserved and uses the learned representations for downstream tasks \cite{hamilton2017representation}. Heterogeneous GRL (HGRL) algorithms can be categorized into three approaches, based on meta path (MPath), message passing (MPass), and relational learning (RL).

\textit{Meta-paths} are paths connected by heterogeneous edges, with flexible length and edge types \cite{sun2011metapath}, e.g., 
Author-Paper-Conference-Paper-Author (APCPA) in citation networks. MPath-based approach transforms given heterogeneous graph into other data structures, e.g., multiple homogeneous graphs \cite{wang2019heterogeneousattention, yun2019graph, zhang2018GraphInception} or sequences of entities \cite{dong2017metapath2vec}, according to the meta-paths to simplify downstream tasks. Both the accuracy and computational cost of downstream tasks can be significantly affected by the number of meta-paths and their specific choices, particularly for high node heterogeneity. However, to leverage edge attributes, e.g., labels, a (much) larger graph needs to be created by converting attributes into additional nodes, making the problem more complex and challenging \cite{yang2020review}.

In contrast, MPass- (e.g., RGCN \cite{schlichtkrull2018modeling}) and RL-based approaches (e.g., DistMult \cite{yang2014embedding}) can leverage node/edge attributes naturally. In MPass-based HGRL \cite{schlichtkrull2018modeling, zitnik2018modeling, wang2019heterogeneousattention}, node attributes are passed as messages along edges, with the edge information determining how messages are aggregated via graph convolution operations \cite{kipf2017gcn}. RL-based HGRL views a heterogeneous graph as a set of triples (i.e., labelled edges) composed of two nodes and their relation (i.e., edge label) and learns a prediction function for such triples and associated attributes. However, both approaches embed all types of nodes/edges into the same vector space, without modelling the semantics of these entities/relations. Some RL-based methods (e.g., \cite{garcia2017kblrn, malone2018knowledge}) design hand-crafted rules to take the semantic information into account but the need for strong prior knowledge limits their applicability to large-scale, complex HGRL problems. Due to the expensive graph convolution operations \cite{ kipf2016variational, kipf2017gcn}, scalability is also a major challenge for MPass-based methods. 

\begin{figure}[t]
	\centering
	\includegraphics[width=\columnwidth]{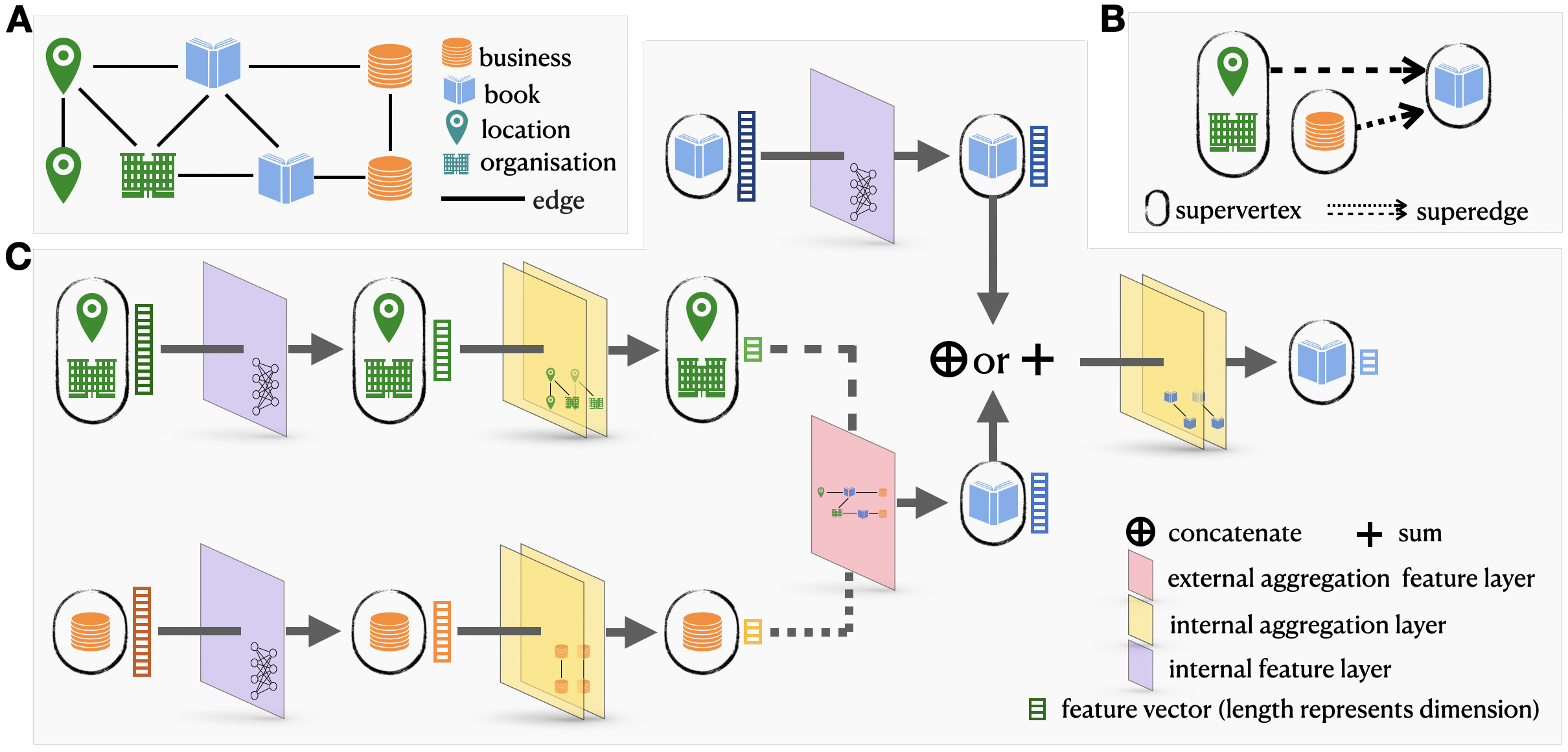}
	\caption{
		GripNet illustration (best viewed in colour). \textbf{A}: A heterogeneous graph $H$ with 4  \textit{types} of nodes (7 nodes in total), which belong to 3 semantic \textit{categories} (one colour for each category), i.e., two types (location and organization, both in green) are of the same category due to their frequent links. 
		\textbf{B}: We segregate the graph in \textbf{A} into three \textit{supervertices} connected by two \textit{directed} \textit{superedges}, forming a directed acyclic \textit{supergraph} $G^S(H)$. The directions of superedges are determined by the \textit{target} of the downstream task, e.g., here we consider the task of book classification so information flows towards the book supervertex. Each supervertex is a subgraph containing nodes of the same category (colour) and edges between them. Each superedge is a \textit{bipartite} subgraph with nodes from two categories (colours) forming two node sets, connected by edges between them. 
		\textbf{C}: GripNet architecture for learning new book representation on the supergraph in \textbf{B}. Information is propagated from the location\&organization (green) and business (orange) supervertices to the book supervertex (blue) via up to three layers. The internal feature layer (purple) reduces the dimension of the original node features. The internal aggregation layer (yellow) aggregates neighbourhood information within the supervertex. The external aggregation feature layer (magenta) aggregates information from all parent supervertices.}
	\label{supergraph}
\end{figure}

The above challenges are mainly due to the rich node attributes, relations, and semantic information in heterogeneous graphs. Our main idea is to \textit{segregate the whole graph into semantically-coherent parts, learn embedding within each part, and pass messages between parts following a task-specific propagation path}. To this end, we propose a new HGRL framework named as \textbf{Gr}aph \textbf{i}nformation \textbf{p}ropagation \textbf{Net}work (GripNet) to leverage the strengths of three existing  approaches:
\begin{itemize} 
\item Firstly, we introduce a novel \textit{supergraph} data structure that segregates a heterogeneous graph into several semantically-coherent subgraphs, named as    \textit{supervertices}, interconnected by heterogeneous bipartite subgraphs, named as \textit{superedges}. By semantically-coherent, we mean that each segregated subgraph contains nodes of the same broad \textit{category} (e.g. gene and protein nodes). We specify the directions of superedges to define the information propagation path between supervertices, in a task-specific manner according to semantic relations and the task(s) of interest. Thus, a supergraph is a directed acyclic graph. Figures \ref{supergraph}A and \ref{supergraph}B show an example.
\item Secondly, the above supergraph structure enables more efficient and effective embedding. We learn new node representations in supervertices sequentially according to the topological ordering of these supervertices defined by the propagation path. This allows us to embed nodes in different supervertices into different embedding spaces, e.g., according to their importance/relevance to the task of interest. This not only offers great modelling flexibility, but also improves scalability by enabling efficient implementation, e.g., by choosing low-dimensional embedding space for less important supervertices. Figure \ref{supergraph}C shows an example.
\end{itemize}
GripNet also provides an efficient solution for data integration. We can construct a supergraph from multiple datasets by modelling each dataset as a supervertex and define their relationships by directed superedges to allow flexible data integration. We evaluate GripNet on link prediction and node classification by constructing seven large-scale complex heterogeneous graph datasets and study its data integration performance. For link prediction, we propose a \textit{categorized negative sampling} strategy to take edge heterogeneity into account. This strategy is generic and can improve the convergence and prediction accuracy of the MPass- and RL-based methods.

\section{Proposed Method: Supergraph and GripNet}
\label{sec:method}

Representation learning for complex heterogeneous graphs is challenging. Existing approaches can not capture semantic relationships and leverage node/edge attributes at the same time, and those graph convolution-based methods are not scalable to large-scale, complex heterogeneous graphs.

Our hypothesis is that because the semantics of nodes can vary greatly for a complex heterogeneous graph, it would be beneficial to learn representations of nodes with large semantic differences separately and then propagate the learned information to serve the need of a particular downstream task of interest. This can lead to different embedding spaces for semantically different nodes, and subsequently improve both predictive accuracy and scalability due to the more compact and effective representations. To realize this idea, we need a new framework to facilitate such separation and learning. Let us consider the following mathematical definition of a heterogeneous graph.

\begin{defn}
	A \emph{heterogeneous} graph $H = (V, E, \mathcal{T}, \mathcal{L}, \tau)$ is a graph with multiple types of nodes and whose edges are labelled. $V$ is the set of nodes and $\mathcal{T}$ is the set of node types. $E = \{(i, j, l)|i,j\in V,\ l\in \mathcal{L}\}$ is the set of labelled edges, where $\mathcal{L}$ is the set of edge labels. The function $\tau : V\rightarrow \mathcal{T}$ is defined as $\tau(i)=t\in \mathcal{T}$ if node $i$ is of type $t$.
\end{defn}
With this definition, we can introduce a novel graph structure, named as \textit{supergraph}, to facilitate our desired separation and learning. 

\subsection{A New Supergraph Model}

Firstly, we define a \textit{categorical partition} of node types $\mathcal{T}=\bigcup_c \mathcal{T}_c$ such that types with each $\mathcal{T}_c$ are \textit{semantically coherent}, i.e., belonging to the same broad \textit{category} $c\in \mathcal{C}$ ($\mathcal{C}$ is the set of all categories). This partition is user defined and there can be a trade-off between prediction performance and memory cost. In our studies, a large $|\mathcal{C}|$ gives better results but also higher computational cost (see the ablation study in Supplementary Material). A naive choice is to assign each type a unique category such that $|\mathcal{C}|=|\mathcal{T}|$, but this is not scalable w.r.t. $|\mathcal{T}|$.

Now, we are ready to define the \emph{supergraph} for a given heterogeneous graph $H$ to describe our proposed information propagation process. To learn node features of different categories in different feature spaces, we segregate the original graph $H$ into several semantically-coherent subgraphs, named as \textit{supervertices},  interconnected by bipartite subgraphs, named as \textit{superedges}. An example of supergraph construction is given in Figure \ref{supergraph}. Formally,

\begin{defn} \textit{Supergraph}, \textit{supervertex}, and \textit{superedge}.
	Given a heterogeneous graph $H = (V, E, \mathcal{T}, \mathcal{L}, \tau)$ and a categorical partition of its node types $\mathcal{T} = \bigcup_c \mathcal{T}_c$, a supervertex $v^S_c$ for a category $c\in\mathcal{C}$ is defined to be the induced subgraph of $H$ from the set of nodes $V_c=\{v\in V | \tau(v)\in c\}$. 
	A superedge $e^S_{cc'}$ connecting two supervertices $v^S_c$ and $v^S_{c'}$ is $H$'s bipartite subgraph $e^S_{cc'}=(V_c, V_{c'}, E_{cc'})$ where $E_{cc'}=\{(v,v', l)\in E | v\in V_c, v'\in V_{c'} \}$, and we say a superedge $e_{cc'}$ exists if $E_{cc'}$ is none-empty. 
	Finally, a supergragh of $H$ is a directed acyclic graph 
	$G^S(H)=(V^S=\{v_c^S|c\in \mathcal{C}\}, E^S=\{e^S_{cc'}|E_{cc'}\neq \varnothing;\ c,c'\in \mathcal{C}\})$.
\end{defn}

The directions of superedges are defined by users according to two criteria: \textbf{1)} the supergraph $G^S$ must be directed acyclic (otherwise we will encounter loopy information propagation) and \textbf{2)} the supervertex on which the ML task of interest (e.g. node classification) is conducted should be a leaf vertex of $G^S$ so that all information can finally be propagated into it. In practice, we typically choose the number of supervertices $|\mathcal{C}|$ to be small for computational efficiency ($\leq 3$ in our experiments).

This new supergraph model is generic and can be used for many problems in heterogeneous graphs, e.g. we can use it as an intuitive and flexible data integration model by constructing a supergraph from multiple datasets with each dataset as a supervertex and their relationships defined by the directions of superedges. In the following, we propose our supergraph-based GripNet for heterogeneous graph representation learning. There can be many ways to realize GripNet. Here we present one simple, basic realization that can be extended to many other architectures by varying individual components. Figure \ref{supergraph}C shows an example of this simple GripNet implementation.

\subsection{Proposed GripNet Architecture}

GripNet learns the embeddings of nodes in supervertices sequentially according to the topological ordering of the supervertices they belong to, which is made possible by designing the supergraph to be a directed acyclic graph \cite{ thulasiraman19925}. For each supervertex $v^S_c$, GripNet has a module for learning node embeddings within $v^S_c$ from the information given by $v^S_c$ and its parent supervertices $\mathcal{N}_c^S=\{v^S_{c'} | e^S_{c'c}\in E^S\}$. Conceptually the module is composed of three layers: 

\textbf{The external aggregation feature layer} of supervertex $v^S_c$ takes the learned node embeddings from its parent supervertices as input, and transform them into features for nodes within $v_c^S$, which are referred to as \textit{external features}. This layer is realized as a set of mutually parallel RGCN encoder sublayers,  one for each parent supervertex $v^S_{c'} \in \mathcal{N}_c^S$. The supervertex $v_{c'}^S$ 's contribution to the external feature of node $i$ in $v^S$ is:
\begin{equation}
\mathbf{r}_{i, \text{ex}}^{c'\rightarrow c} = \sum_{l\in \mathcal{L}_{c'c}} \sum_{ j\in \mathcal{N}_{i,l}^{cc'}}  \dfrac{1}{|\mathcal{N}_{i,l}^{cc'}|} \mathbf{W}_{l}^{(\text{ex})} \mathbf{z}_{j}^{c'}, 
\end{equation}
where $\mathbf{z}^c_j$ is the embedding of node $j$ in $v^S_{c'}$, $\mathcal{L}_{c'c}$ denotes the set of possible edge labels in superedge $e_{cc'}^S$ (a bipartite subgraph), $\mathcal{N}_{i,l}^{cc'}$ is the set of node $i$'s neighbours in $e_{cc'}^S$ linked by an edge of type $l$, and $\mathbf{W}^{(\text{ex})}$s are learnable parameters. The total external feature of node $i$ in $v^S_c$ is obtained by summing over all supervertices' contributions and applying an activation function, such as ReLU:
\begin{equation}
\mathbf{r}^{c}_{i, \text{ex}} = \text{ReLU} \left( \dfrac{1}{|\mathcal{N}_c^S| }\sum_{c' \in \mathcal{N}_c^S}  \mathbf{r}_{i, \text{ex}}^{c'\rightarrow c} \right). 
\end{equation}
Note that if a supervertex doesn't have any parent supervertex, i.e. if it's a root vertex in the supergraph, then the external feature layer becomes degenerated   and we can simply choose the external feature vectors as zero vectors: $\mathbf{r}^{c}_{i, \text{ex}} = \mathbf{0}$.

\textbf{The internal feature layer} of supervertex $v^S_c$ maps the original node features, typically one-hot vectors, into the same space that external features $\{\mathbf{r}_{i,\text{ex}}^c\}$ live in. This layer is constructed as a linear transformation $\mathbf{W}^{(\text{in})}$ followed by a nonlinear activation function, such as ReLU. Its output on node $i$  in $v^S_c$, referred to as \textit{internal features},  is:
\begin{equation}
\mathbf{r}^c_{i, \text{in}} = \text{ReLU}(\mathbf{W}^{(\text{in})} \mathbf{x}_{i}^c),
\end{equation}
where $\mathbf{x}^c_{i}$ is the original feature vector of node $i$. We combine, the internal and external features to obtain the \textit{total feature} of node $i$, e.g. by concatenation or summation as 
\begin{equation}
\mathbf{r}^c_i = ( \mathbf{r}^c_{i, \text{ex}} \oplus \mathbf{r}^c_{i, \text{in}}) \:\: \mathrm{or} \:\: (\mathbf{r}^c_{i, \text{ex}} + \mathbf{r}^c_{i, \text{in}}),
\end{equation}
where $\oplus$ denotes concatenation and $+$ denotes summation. We set the default in GripNet to be concatenation $\oplus$, with additional studies in Supplementary Material.

\textbf{The internal aggregation layer} \label{sec:up} takes the total features $\{ \mathbf{r}^c_{i}\}$ as the input $\mathbf{u}_i^0 = \mathbf{r}_i^c$, update them within the supervertex $v_c^S$, and obtain the final embeddings for nodes in $v_c^S$. This layer is designed to be composed of $n$ RGCN sublayers concatenated together, and the update rule between the $m$th and the $(m+1)$th sublayer is:
\begin{equation}
\mathbf{u}_{i}^{m+1} =  \mathbf{Re}\mathbf{LU} \left(
\sum_{l\in \mathcal{L}_c, j \in \mathcal{N}_{i,l}^{c}} 
\dfrac{1}{\mathcal{N}_{i,l}^{c}} \mathbf{W}_{l, m}^{(\mathbf{ia})} \mathbf{u}_{j}^m
+ \mathbf{W}_{0, m}^{(\mathbf{ia})} \mathbf{u}_{i}^m \right)
\label{eq: u}
\end{equation}
where $\mathcal{L}_c$ is the set of possible edge labels in $v_c^S$,  $\mathcal{N}_{i,l}^{c}$ denotes the neighbours of node $i$ linked by edges of type $l$, and $\mathbf{W}$s are learnable parameters. 
The output of the last RGCN layer $\mathbf{z}^c_i = \mathbf{u}^n_i$ is the final learned embedding of node $i$ in $v_c^S$. These embeddings can then be used for computing external features of $v_c^S$ 's child supervertices or downstream ML tasks.

Here we consider two popular tasks: link prediction and  node classification. We follow the popular encoder-decoder framework in graph representation learning \cite{hamilton2017representation}. The GripNet representation learning introduced above learns a new vectorial embedding for each node, which is the GripNet encoder. For downstream tasks, the learned GripNet representation is fed into a GripNet decoder to infer graph properties, such as the existence probabilities of node attributes and edges.

\subsection{Link Prediction}
We focus on the case where the edges to be predicted are all located in a single supervertex $v_c^S$. The method presented below can be slightly modified to predict edges located in a superedge. We formulate link prediction as a binary classification of exist vs non-exist and adopt the DistMult  factorization \cite{yang2014embedding} as the GripNet decoder for link prediction. Given nodes $i$ and  $j$ in supervertex $v_c^S$ and a possible edge label $l\in \mathcal{L}_{c}$, the probability that the edge $(i, j, l)$ exists is computed as:
\begin{equation}
f(i, j, l)  = \text{sigmod}((\mathbf{z}_i^c)^{\top} \mathbf{M}_l (\mathbf{z}^{c}_j)),
\end{equation}
where $\mathbf{M}_l$ is a trainable diagonal matrix associated with the edge label $l$, and $\mathbf{z}_i$, $\mathbf{z}_i$ are the embeddings for nodes $i$, $j$ . To train the model, we need both positive and negative instances of edges. Positive edges are the observed edges $E_c$, while negative ones requires sampling. Traditional negative sampling method  \cite{mikolov2013distributed, schlichtkrull2018modeling, yang2014embedding, zitnik2018modeling} does not take edge heterogeneity of graph into account. Thus, we propose a \emph{categorized negative sampling} strategy for GripNet decoder in link prediction:

\textbf{Categorized negative sampling (CNS)}. To sample a negative edge with label $l$ in supervertex $v_c^S=(V_c, E_c)$, we propose a sampler to randomly choose a source-target node pair from the set $\{(i, j)| i, j \in V_c, \ (i, j, l)\notin E_c\}$. In contrast, in previous methods negative edges are chosen from the whole heterogeneous graph, and are constructed by corrupting (redirecting) one end of each positive instance. For each edge label $l$, we enforce the number of negative samples to be the same as that of positive instances in $v_c^S$.

We use the \textit{cross-entropy loss} to optimize the model, aiming to assign higher probabilities to observed edges and lower probabilities to undiscovered ones, 
\begin{equation}
L^{\text{(lp)}} = - \sum_{(i,j, l) \in E_c} \log f(i,j,l)  - \sum_{(i,j, l) \in {N}(E_c)} \log(1-f(i,j, l)), 
\end{equation}
where $E_c$ is the set of positive edges in $v_c^S$, and ${N}(E_c)$ is the set of negative edges sampled by categorized negative sampler according to $E_c$.

\subsection{Node Classification}
In node classification, we consider the task of predicting a specific node attribute with possible values from a set $\mathcal{A}$ for nodes in a given supervertex $v_c^S$ (their types or categories are assumed to be known). The task can be viewed as a multiclass classification problem. Our GripNet decoder for node classification applies a linear transformation $\mathbf{W}^{\text{(nc)}}$ followed by a softmax activation function to $i^{th}$ node's embedding vector $\mathbf{z}_i^c$ to obtain the probability distribution vector $\mathbf{p}_i$ over $\mathcal{A}$:
\begin{equation}
\mathbf{p}_i = \text{softmax}(\mathbf{W}^{\text{(nc)}}\mathbf{z}_i^c).
\end{equation}
We minimize the following \textit{cross-entropy} loss during training:
\begin{equation}
L^{\text{(nc)}} =  -\sum_{i\in V_c,\   a\in \mathcal{A}} s_{ia} \ln(p_{ia}), 
\end{equation}
where $\mathbf{s}$ is the one hot representation of true attributes:
$s_{ia}=\mathbbm{1} [\text{node }i\text{'s true attribute is } a]$, 
$V_c$ is the set of nodes in supervertex $v_c^S$ and $p_{ia}$ is the $a$th  component of $\mathbf{p}_i$. 

\subsection{Relationship with Other Works}
GripNet is conceptually related to previous MPath-based \cite{dong2017metapath2vec, wang2019heterogeneousattention, yun2019graph, zhang2018GraphInception}, MPass-based \cite{schlichtkrull2018modeling, zitnik2018modeling}  and RL-based \cite{garcia2017kblrn, malone2018knowledge, yang2014embedding} heterogeneous graph representation learning approaches mentioned in the Introduction. GripNet and these methods all follow a popular encoder-decoder framework for graph representation learning \cite{hamilton2017representation}.  
GripNet parameters are learned in an end-to-end manner as in many other works \cite{schlichtkrull2018modeling, wang2019heterogeneousattention, yun2019graph, zitnik2018modeling}.

Our GripNet \textit{encoder} is closely related to RL-based HGRL with GCNs \cite{schlichtkrull2018modeling, zitnik2018modeling}. Besides, the GripNet link prediction decoder relies on DistMult \cite{yang2014embedding}, which is a simpler and more effective case of the RESCAL factorization \cite{nickel2011three}. In principle, this decoder can be replaced with numerous alternative tensor factorization methods or scoring functions \cite{bordes2013transE, chang2014typed, socher2013reasoning, trouillon2016complex} for RL.

\section{Experiments}
\label{sec:experiments}

We evaluate GripNet on \textit{large-scale}, complex heterogeneous graphs below. All scripts, datasets and models are described and open-sourced online at \url{https://github.com/NYXFLOWER/GripNet}. Additional experimental settings required for conducting experiments are included in the Supplementary Material.

\begin{table}[t]
	\vspace{-4mm}
	\caption{Dataset Statistics. $|V|$: $\#$nodes, $|E|$: $\#$edges, $|V_{p}|$ ($|E_{p}|$): $\#$node ($\#$edge) to be predicted, $|\mathcal{C}_p|$: $\#$candidate labels, $|\mathcal{T}|$: $\#$node types, $|\mathcal{L}|$: $\#$edge labels, \textbf{Items}: average $\#$items to be predicted per candidate label. Node types contains in datasets on drug(d), gene(g), paper(p), author(a), book(bo), business(bu), organization(o) and location(l).}
	\label{tab:datasets}
	\centering
	\scalebox{0.9}{
	\begin{tabular}{crrrrrrrrc}
		\toprule
		\textbf{Dataset}& $|V|$   & $|\mathcal{T}|$ & $|E|$   & $|\mathcal{L}|$ & $|V_p|$ & $|E_p|$ & $|\mathcal{C}_p|$ & \textbf{Items} & \textbf{Node types}\\
		\midrule
		PoSE-0& 4,285& 2& 4,725,690& 1,100& -& 4,625,608& 1,097 & 4,217 & d, g\\
		PoSE-1& 19,365& 2& 2,621,423& 864& -& 1,171,603& 861 & 1,361 & d, g\\
		PoSE-2& 19,726& 2& 6,075,428& 1,100& -& 4,625,608&  1,097 & 4,217 & d, g\\
		\midrule
		Aminer& 397,477& 2& 1,265,593& 3& 124,806& -&  7& 17,829& a, p\\
		\midrule
		Freebase-a& 14,989& 1& 12,556& 1& 14,989& -& 8 & 1,873& bo\\
		Freebase-b& 354,961& 2& 848,032& 3& 14,989& -& 8 & 1,873& bo, bu\\
		Freebase-c& 457,504& 3& 998,663& 5& 14,989& -& 8 & 1,873& bo, bu, o\\
		Freebase-d& 1,100,400& 4& 3,354,079& 8& 14,989& -& 8 & 1,873& bo, bu, o, l\\
		\bottomrule
	\end{tabular}
	 }
\end{table}

\textbf{Dataset.} 
For link prediction with a large number of different relations, we consider the BioSNAP-Decagon dataset with 6M edges and 1K different edge labels in total \cite{zitnik2018modeling} by constructing a series of three sub-datasets named as \textbf{PoSE-0}, \textbf{-1}, and \textbf{-2}, with varying proportions of task-related nodes. For node classification with a large number of labelled nodes, we construct the \textbf{Aminer} dataset with 124K labelled nodes and the Freebase-series datasets with four sub-datasets, \textbf{Freebase-a}, \textbf{-b}, \textbf{-c} and \textbf{-d}, with increasing scale and heterogeneity. These datasets are briefly described below, with additional details in Supplementary Material. Table \ref{tab:datasets} reports their statistics.
\begin{itemize}
	\item The \textbf{PoSE-series} are three sub-datasets from the BioSNAP-Decagon  dataset \cite{snap} for \textbf{Po}lypharmacy \textbf{S}ide \textbf{E}ffect (POSE) prediction \cite{zitnik2018modeling}: PP, GhG and ChChSe contain relations between genes, gene and drug, and drugs respectively. We integrate the datasets by GeneID \cite{maglott2005entrez} and PubChem CID \cite{kim2016pubchem}. The key difference among PoSE-0, 1 and 2 is the number of gene/drug nodes that are not directly linked with any drug/gene.
	\item \textbf{Aminer} is constructed from the public Aminer Academic Network dataset \cite{Tang:08KDD}. Following the top venue classification in Google Scholar\footnote{\url{https://scholar.google.com/citations?view_op=top_venues&hl=en&vq=eng_enggeneral}}, we select seven popular categories in computer science (CS) and 20 top venues for each category. Following \cite{dong2017metapath2vec}, we assign each author node a CS-category label with the most publications.
	\item The \textbf{Freebase-series} are subsets of Freebase dataset \cite{yang2020review} with books classified into 8 categories. As the last column of Table \ref{tab:datasets} shown, Freebase-a contains book nodes and their relations. Business, organization and location information are integrated cumulatively to construct the Freebase-b, -c and -d datasets respectively.
\end{itemize}

\textbf{Evaluation Metrics.}
We use the area under the precision-recall curve (AUPRC), the receiver-operating characteristic (AUROC), and average precision at 50 (AP@50) to evaluate link prediction performance, for each edge label first and then taking their average. We use the micro-averaged F1 scores (Micro-F1) and macro-averaged F1 scores (Macro-F1) to evaluate node classification. We also evaluate the GPU memory usage during model training uses tools in the 
\textit{pytorch\_memlab}  package\footnote{\url{https://github.com/Stonesjtu/pytorch_memlab}}, and  the computational time.

\textbf{Methods Compared.}
In the PoSE link prediction task, \textbf{DECAGON} \cite{zitnik2018modeling} is a pioneering work but with high computational requirement that we were not able to satisfy. A more recent work \cite{malone2018knowledge} showed \textbf{DistMult} \cite{yang2014embedding} has better performance and lower computational cost than DECAGON so we choose it for comparison. Three KG embedding models \textbf{TransE} \cite{bordes2013transE}, \textbf{ComplEx} \cite{trouillon2016complex} and \textbf{RotatE} \cite{sun2019rotate} are compared as they are considered as baselines for multi-relation link prediction by Open Graph Benchmark \cite{hu2020ogb}. We also compare with \textbf{RGCN} \cite{schlichtkrull2018modeling}, which shows good performance on standard datasets for heterogeneous graphs. In node classification, we choose one MPass/RL-based method \textbf{RGCN} \cite{schlichtkrull2018modeling} and two popular GRL methods \textbf{GCN} \cite{kipf2017gcn} and \textbf{GAT} \cite{velickovic2018graph}.

\subsection{Experimental Configuration} 
We train and test all considered methods on NVIDIA GV100GL (Tesla V100 PCIe 32 GB VRAM). For fair comparison and to avoid costly tuning on large-scale data, we use the following configurations for all methods: 1) Set the learnable embeddings' dimension for the nodes of interest to $32$ in node classification, 2) Choose the embedding dimension for each model such that training can be performed on the GPU memory budget \cite{hu2020ogb} in link prediction, 3) initialize weights using \textit{Xavier initialization} \cite{glorot2010understanding} , 4) optimize models end-to-end by full-batch with the \textit{Adam optimizer} \cite{adam} with learning rate $0.01$, and 5) train a fixed 100 epochs for all the experiments. 

In addition, when implementing methods involving graph convolution operations, an additional embedding layer with trainable parameters $\mathbf{W} \in \mathbb{R}^{|\mathcal{V}| \times 256}$ is added before convolution layers for computational efficiency, as the datasets contain up to millions of nodes. We do a stratified split into $90\%$ for training and $10\%$ for testing for both link prediction and node classification. With the above configuration, we did construct validation sets to tune hyper-parameters for performance optimization. Each experiment is repeated 10 times with different random seeds, and all test results are reported with the mean and unbiased standard deviation.

\textbf{Implementation.} We implement GripNet models with \textit{PyTorch} \cite{torch} and \textit{PyTorch-Geometric} packages \cite{pytorch}. We use the model implementation of GCN, RGCN and GAT model provided by \textit{PyTorch-Geometric} \cite{pytorch}. The implementation of TransE , RotatE , ComplEx  and DistMult  model are provided by \textit{Open Graph Benchmark} \cite{hu2020ogb}. Due to the large-scale data size, we improved the RGCN implementation to reduce its memory requirements for the link prediction task. 

\textbf{Supergraph Construction.} Figure \ref{supergraph} gives an example on supergraph construction for book classification on Freebase-d. The supergraph construction for Freebase-c is similar that for Freebase-d. For datasets with only two node types,  PoSE-series, Aminer, and Freebase-b, their supergraph only contains two supervertices and the superedge is directed to the task-related supervertex from the other. According to the PoSE-series supergraphs, we have two GripNet implementations: \textbf{GripNet-l} and \textbf{GripNet-r}, which focus on capturing information on the leaf and root supervertex respectively. Normally, the internal aggregation layer (see Eq. \ref{eq: u}) on each supervertex contains a RGCN sublayer, while we use two-sublayer internal aggregation layer on the leaf supervertex for GripNet-l and on the root supervertex for GripNet-r.

\begin{table}[t]
	\caption{Multi-relational Link prediction results in \textbf{AUROC}. The best result is in bold and second best one is underlined. Analogous trends hold for AUPRC and AP@50. \textbf{Dim}: the embedding dimension for each model such that the training peak GPU memory usage is around 30GB. \textbf{TpE}: computational time per epoch (including training and testing score computation)
	}
	\label{tab:mrlp}
	\centering
	\scalebox{0.9}{
		\begin{tabular}{lrrrrrrrrr}
			\toprule
			& \multicolumn{3}{c}{PoSE-0} & \multicolumn{3}{c}{PoSE-1}  & \multicolumn{3}{c}{PoSE-2}  \\
			\cmidrule(r){2-4} \cmidrule(r){5-7} \cmidrule(r){8-10}
			Model     & \textbf{AUROC}    & \textbf{Dim}    & \textbf{TpE(s)} & \textbf{AUROC}     & \textbf{Dim} & \textbf{TpE(s)} & \textbf{AUROC}     & \textbf{Dim} & \textbf{TpE(s)}\\
			\midrule
			TransE & $0.718 \pm 0.006$ & $134$ & $30.0$ & $0.826 \pm 0.007$ & $170$ & $21.5$ & $0.751 \pm 0.006$ & $106$ & $30.8$ \\
			RotatE & $0.891 \pm 0.003$ & $32$ & $29.7$ & $0.857 \pm 0.006$ & $38$ & $20.7$ &  $0.851 \pm 0.004$ & $24$ & $30.6$ \\
			ComplEx & $0.595 \pm 0.004$  & $34$ & $\underline{29.2}$ & $0.844 \pm 0.003$ & $48$ & $\mathbf{20.3}$ & $0.560 \pm 0.004$ & $26$ & $\underline{30.4}$ \\
			DistMult  & $0.843 \pm 0.008$ & $84$ & $\mathbf{29.0}$ & $0.859 \pm 0.008$ & $108$ & $\underline{20.4}$ & $0.877 \pm 0.009$ & $66$ & $\mathbf{29.7}$ \\
			RGCN  & $\underline{0.905 \pm 0.006}$ & $76$ & $47.5$ & $0.909 \pm 0.005$ & $94$ & $37.1$ & $0.850 \pm 0.003$ & $58$ & $60.3$ \\
			\midrule
			\textbf{GripNet-l} & $\mathbf{0.920 \pm 0.003}$ & $32$ & $39.9$ & $\mathbf{0.914 \pm 0.006}$ & $120$ & $24.2$ & $\underline{0.918 \pm 0.004}$ & $32$ & $40.3$ \\
			\textbf{GripNet-r} & $\mathbf{0.920 \pm 0.003}$ & $16$ & $41.1$ & $\underline{0.912 \pm 0.003}$ & $58$ & $25.0$ & $\mathbf{0.920 \pm 0.050}$ & $16$ & $41.4$ \\
			\bottomrule
	\end{tabular}}
\end{table}

\begin{table}[t]
	\caption{Node classification results in \textbf{Micro-F1}. The best result is in bold and second best one is underlined. Analogous trends hold for Macro-F1. \textbf{TpE} are same as those in Table \ref{tab:mrlp}.
	}
	\centering
	\scalebox{0.9}{
		\centering
		\begin{tabular}{lrrrrrrrr}
			\toprule
			& \multicolumn{2}{c}{Aminer} & \multicolumn{2}{c}{Freebase-b}  & \multicolumn{2}{c}{Freebase-c} & \multicolumn{2}{c}{Freebase-d} \\
			\cmidrule(r){2-3} \cmidrule(r){4-5} \cmidrule(r){6-7} \cmidrule(r){8-9}
			Model     & \textbf{Mi-F1}      & \textbf{TpE(s)}    & \textbf{Mi-F1}  & \textbf{TpE(s)}      & \textbf{Mi-F1}  & \textbf{TpE(s)}  & \textbf{Mi-F1}  & \textbf{TpE(s)}   \\
			\midrule
			GCN  & $\underline{0.907 \pm 0.010} $&  $\underline{0.26}$& $0.433 \pm 0.010$  & $\underline{0.17}$& $\underline{0.508 \pm 0.011}$& $\underline{0.22}$& $0.563 \pm 0.009$ & $\underline{0.51}$\\
			
			GAT& $0.899\pm 0.003$&  $0.36$& $\underline{0.454\pm 0.005}$&  $0.43$& $0.498\pm 0.004$&  $0.50$& $\underline{0.564\pm 0.004}$ &$0.94$\\
			RGCN  & $0.882\pm 0.005$&  $2.62$& $0.365\pm 0.005$&  $1.09$& $0.464\pm 0.006$&  $1.31$& $0.506\pm 0.006$&  $2.67$\\
			\midrule
			\textbf{GripNet} & $\mathbf{0.920 \pm 0.002}$  & $\mathbf{0.25}$& $\mathbf{0.464\pm 0.002}$  & $\mathbf{0.08}$ & $\mathbf{0.564\pm 0.002}$ &  $\mathbf{0.16}$& $\mathbf{0.592\pm 0.002}$ &$\mathbf{0.36}$\\
			\bottomrule
	\end{tabular}}
	\label{tab:ncm}
\end{table}

\subsection{Performance Comparison}
Tables \ref{tab:mrlp} and \ref{tab:ncm} report the experimental results in terms of accuracy (AUROC / Micro-F1), time (training and testing time per epoch) for link prediction and node classification respectively, with the best result in bold and the second-best underlined. Additional evaluations on the GPU peak memory usage for each method are provided in the Supplementary Material. With effective information propagation between supervertices, GripNet achieves the best AUROC/Micro-F1 scores on all seven datasets.

\begin{figure}[t]
	\centering
	\includegraphics[width=\columnwidth]{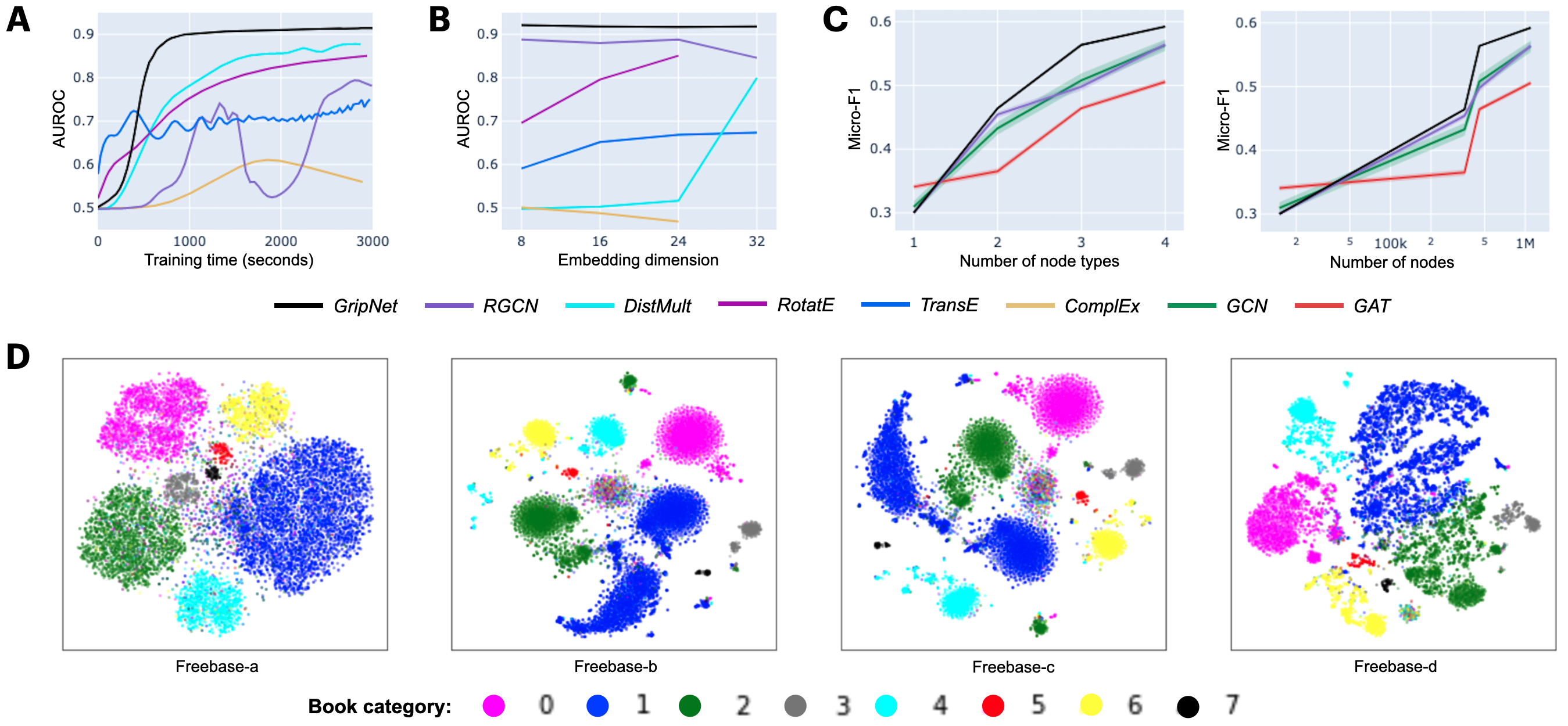}
	\caption{\textbf{A}: AUROC over training time for different models on PoSE-2.
		\textbf{B}: AUROC at different embedding dimensions on PoSE-2. \textbf{C}: Micro-F1 on the four Freebase datasets plotted with respect to the number of node types (graph heterogeneity) and the number of nodes, with standard deviations as shades (only visible for GCN). \textbf{D}: Visualization of GripNet embedding for eight book categories in  Freebase-series datasets using t-SNE \cite{van2014accelerating} to study the data integration performance. Colours represent different book categories.}
	\label{fig:result}
\end{figure}

From the link prediction results in Table \ref{tab:mrlp} and Figure \ref{fig:result}B,  GripNet has the best prediction accuracy under a maximum GPU memory of 30GB and varying embedding dimensions, and baseline models are more sensitive to embedding dimension than GripNet. From Figure \ref{fig:result}B, the AUROC of GripNet does not increase with increasing embedding dimension, while at the lowest embedding dimension 8, GripNet greatly outperformed the RL-based methods (TransE, RotatE, ComplEx and DistMult). Although RL-based methods have a shorter per-epoch running time than GripNet (from Table \ref{tab:mrlp}), GripNet converges much faster (from Figure \ref{fig:result}A) and fewer epochs are needed in practice.

In node classification (Table \ref{tab:ncm}), the F1 scores for different models have small difference on Aminer but more variance for the Freebase-series. This indicates that the information between supervertices of Aminer is less complementary and GripNet leads to more improvement when the dataset heterogeneity increases. In addition, compared with other MPass-based methods (GCN, RGCN, and GAT), GripNet is much faster (from Table \ref{tab:ncm})  and consumes much less GPU memory (see Supplementary Material) because of its segregated architecture that reduces unnecessary expensive graph convolution operations.

\subsection{Data Integration Studies}
Heterogeneous graphs are useful for integrating multiple datasets. Thus, GripNet can be used for data integration, where each supervertex represents a dataset and superedges define the relationships between datasets as information propagation paths. For example, Freebase-b has two supervertices (book and business). To add/integrate an ``organization dataset'' into Freebase-b, we add a supervertex of organization first, and then create a superedge from the organization supervertex to the book supervertex because of the close association between organization and book (based on existing knowledge).

In Figure \ref{fig:result}C, Freebase-a contains only one node type (book) while Freebase-d contains four node types because it has integrated three additional types of data (business, organization, and location). From the figure, we can see that GripNet has the biggest improvement in book classification accuracy by 99\% on Freebase-d over on Freebase-a, while  GCN, GAT, and RGCN have a smaller improvement of 86\%, 65\%, and 69\%, respectively. Figure \ref{fig:result}C also shows that the improvement by GripNet increases as the heterogeneity of the graph increases. In Fig. \ref{fig:result}D, we visualize book embedding learned by GripNet in the four Freebase-series datasets to show that as we integrate data of different types, the boundaries between book categories become clearer.

\section{Conclusion}
This paper proposed a new supergraph data structure and a new Graph Information Propagation (GripNet) framework for learning node representations on heterogeneous graphs. We evaluated GripNet on seven large-scale datasets in  link prediction, node classification, and data integration. GripNet achieved the best overall performance on these datasets.

\bibliographystyle{plain}
\bibliography{references}

\begin{thebibliography}{10}

\bibitem{bordes2013transE}
Antoine Bordes, Nicolas Usunier, Alberto Garcia-Duran, Jason Weston, and Oksana
  Yakhnenko.
\newblock Translating embeddings for modeling multi-relational data.
\newblock In {\em Advances in neural information processing systems}, pages
  2787--2795, 2013.

\bibitem{chang2014typed}
Kai-Wei Chang, Wen-tau Yih, Bishan Yang, and Christopher Meek.
\newblock Typed tensor decomposition of knowledge bases for relation
  extraction.
\newblock In {\em Proceedings of the 2014 Conference on Empirical Methods in
  Natural Language Processing (EMNLP)}, pages 1568--1579, 2014.

\bibitem{dong2017metapath2vec}
Yuxiao Dong, Nitesh~V Chawla, and Ananthram Swami.
\newblock metapath2vec: Scalable representation learning for heterogeneous
  networks.
\newblock In {\em Proceedings of the 23rd ACM SIGKDD international conference
  on knowledge discovery and data mining}, pages 135--144, 2017.

\bibitem{pytorch}
Matthias Fey and Jan~E. Lenssen.
\newblock Fast graph representation learning with {PyTorch Geometric}.
\newblock {\em ICLR Workshop on Representation Learning on Graphs and
  Manifolds}, 2019.

\bibitem{garcia2017kblrn}
Alberto Garcia-Duran and Mathias Niepert.
\newblock Kblrn: End-to-end learning of knowledge base representations with
  latent, relational, and numerical features.
\newblock {\em arXiv preprint arXiv:1709.04676}, 2017.

\bibitem{gilmer2017message}
Justin Gilmer, Samuel~S Schoenholz, Patrick~F Riley, Oriol Vinyals, and
  George~E Dahl.
\newblock Neural message passing for quantum chemistry.
\newblock In {\em Proceedings of the 34th International Conference on Machine
  Learning-Volume 70}, pages 1263--1272. JMLR. org, 2017.

\bibitem{glorot2010understanding}
Xavier Glorot and Yoshua Bengio.
\newblock Understanding the difficulty of training deep feedforward neural
  networks.
\newblock In {\em Proceedings of the thirteenth international conference on
  artificial intelligence and statistics}, pages 249--256, 2010.

\bibitem{hamilton2017representation}
William~L Hamilton, Rex Ying, and Jure Leskovec.
\newblock Representation learning on graphs: Methods and applications.
\newblock {\em arXiv preprint arXiv:1709.05584}, 2017.

\bibitem{kim2016pubchem}
Sunghwan Kim, Paul~A Thiessen, Evan~E Bolton, Jie Chen, Gang Fu, Asta
  Gindulyte, Lianyi Han, Jane He, Siqian He, Benjamin~A Shoemaker, et~al.
\newblock Pubchem substance and compound databases.
\newblock {\em Nucleic acids research}, 44(D1):D1202--D1213, 2016.

\bibitem{adam}
Diederik~P Kingma and Jimmy Ba.
\newblock Adam: A method for stochastic optimization.
\newblock {\em arXiv preprint arXiv:1412.6980}, 2014.

\bibitem{kipf2016variational}
Thomas~N Kipf and Max Welling.
\newblock Variational graph auto-encoders.
\newblock {\em In NIPS Workshop on Bayesian Deep Learning}, 2016.

\bibitem{kipf2017gcn}
Thomas~N Kipf and Max Welling.
\newblock Semi-supervised classification with graph convolutional networks.
\newblock {\em International Conference on Learning Representations}, 2017.

\bibitem{maglott2005entrez}
Donna Maglott, Jim Ostell, Kim~D Pruitt, and Tatiana Tatusova.
\newblock Entrez gene: gene-centered information at ncbi.
\newblock {\em Nucleic acids research}, 33(suppl\_1):D54--D58, 2005.

\bibitem{malone2018knowledge}
Brandon Malone, Alberto Garc{\'\i}a-Dur{\'a}n, and Mathias Niepert.
\newblock Knowledge graph completion to predict polypharmacy side effects.
\newblock In {\em International Conference on Data Integration in the Life
  Sciences}, pages 144--149. Springer, 2018.

\bibitem{snap}
Sagar~Maheshwari Marinka~Zitnik, Rok~Sosic and Jure Leskovec.
\newblock {BioSNAP Datasets}: {Stanford} biomedical network dataset collection.
\newblock \url{http://snap.stanford.edu/biodata}, August 2018.

\bibitem{mikolov2013distributed}
Tomas Mikolov, Ilya Sutskever, Kai Chen, Greg~S Corrado, and Jeff Dean.
\newblock Distributed representations of words and phrases and their
  compositionality.
\newblock In {\em Advances in neural information processing systems}, pages
  3111--3119, 2013.

\bibitem{nickel2011three}
Maximilian Nickel, Volker Tresp, and Hans-Peter Kriegel.
\newblock A three-way model for collective learning on multi-relational data.
\newblock In {\em Icml}, volume~11, pages 809--816, 2011.

\bibitem{torch}
Adam Paszke, Sam Gross, Soumith Chintala, Gregory Chanan, Edward Yang, Zachary
  DeVito, Zeming Lin, Alban Desmaison, Luca Antiga, and Adam Lerer.
\newblock Automatic differentiation in {PyTorch}.
\newblock In {\em NIPS Autodiff Workshop}, 2017.

\bibitem{schlichtkrull2018modeling}
Michael Schlichtkrull, Thomas~N Kipf, Peter Bloem, Rianne Van Den~Berg, Ivan
  Titov, and Max Welling.
\newblock Modeling relational data with graph convolutional networks.
\newblock In {\em European Semantic Web Conference}, pages 593--607. Springer,
  2018.

\bibitem{socher2013reasoning}
Richard Socher, Danqi Chen, Christopher~D Manning, and Andrew Ng.
\newblock Reasoning with neural tensor networks for knowledge base completion.
\newblock In {\em Advances in neural information processing systems}, pages
  926--934, 2013.

\bibitem{sun2019vgraph}
Fan-Yun Sun, Meng Qu, Jordan Hoffmann, Chin-Wei Huang, and Jian Tang.
\newblock vgraph: A generative model for joint community detection and node
  representation learning.
\newblock In {\em Advances in Neural Information Processing Systems}, pages
  512--522, 2019.

\bibitem{sun2011metapath}
Yizhou Sun, Jiawei Han, Xifeng Yan, Philip~S Yu, and Tianyi Wu.
\newblock Pathsim: Meta path-based top-k similarity search in heterogeneous
  information networks.
\newblock {\em Proceedings of the VLDB Endowment}, 4(11):992--1003, 2011.

\bibitem{sun2019rotate}
Zhiqing Sun, Zhi-Hong Deng, Jian-Yun Nie, and Jian Tang.
\newblock Rotate: Knowledge graph embedding by relational rotation in complex
  space.
\newblock International Conference on Machine Learning (ICML), 2019.

\bibitem{Tang:08KDD}
Jie Tang, Jing Zhang, Limin Yao, Juanzi Li, Li~Zhang, and Zhong Su.
\newblock Arnetminer: Extraction and mining of academic social networks.
\newblock In {\em KDD'08}, pages 990--998, 2008.

\bibitem{thulasiraman19925}
K~Thulasiraman and MNS Swamy.
\newblock Acyclic directed graphs.
\newblock In John Wiley and Son, editors, {\em Graphs: Theory and Algorithms},
  chapter 5.7, page 118. 1992.

\bibitem{trouillon2016complex}
Th{\'e}o Trouillon, Johannes Welbl, Sebastian Riedel, {\'E}ric Gaussier, and
  Guillaume Bouchard.
\newblock Complex embeddings for simple link prediction.
\newblock International Conference on Machine Learning (ICML), 2016.

\bibitem{van2014accelerating}
Laurens Van Der~Maaten.
\newblock Accelerating t-sne using tree-based algorithms.
\newblock {\em The Journal of Machine Learning Research}, 15(1):3221--3245,
  2014.

\bibitem{velickovic2018graph}
Petar Veli{\v{c}}kovi{\'{c}}, Guillem Cucurull, Arantxa Casanova, Adriana
  Romero, Pietro Li{\`{o}}, and Yoshua Bengio.
\newblock {Graph Attention Networks}.
\newblock {\em International Conference on Learning Representations}, 2018.
\newblock accepted as poster.

\bibitem{wang2019heterogeneousattention}
Xiao Wang, Houye Ji, Chuan Shi, Bai Wang, Yanfang Ye, Peng Cui, and Philip~S
  Yu.
\newblock Heterogeneous graph attention network.
\newblock In {\em The World Wide Web Conference}, pages 2022--2032, 2019.

\bibitem{hu2020ogb}
Marinka Zitnik Yuxiao Dong Hongyu Ren Bowen Liu Michele Catasta Jure~Leskovec
  Weihua~Hu, Matthias~Fey.
\newblock Open graph benchmark: Datasets for machine learning on graphs.
\newblock {\em arXiv preprint arXiv:2005.00687}, 2020.

\bibitem{wu2019graph}
Yongji Wu, Defu Lian, Shuowei Jin, and Enhong Chen.
\newblock Graph convolutional networks on user mobility heterogeneous graphs
  for social relationship inference.
\newblock In {\em Proceedings of the Twenty-Eighth International Joint
  Conference on Artificial Intelligence}, 2019.

\bibitem{yang2014embedding}
Bishan Yang, Wen-tau Yih, Xiaodong He, Jianfeng Gao, and Li~Deng.
\newblock Embedding entities and relations for learning and inference in
  knowledge bases.
\newblock {\em arXiv preprint arXiv:1412.6575}, 2014.

\bibitem{yang2020review}
Carl Yang, Yuxin Xiao, Yu~Zhang, Yizhou Sun, and Jiawei Han.
\newblock Heterogeneous network representation learning: Survey, benchmark,
  evaluation, and beyond.
\newblock {\em arXiv preprint arXiv:2004.00216}, 2020.

\bibitem{yun2019graph}
Seongjun Yun, Minbyul Jeong, Raehyun Kim, Jaewoo Kang, and Hyunwoo~J Kim.
\newblock Graph transformer networks.
\newblock In {\em Advances in Neural Information Processing Systems}, pages
  11960--11970, 2019.

\bibitem{zhang2015geosoca}
Jia-Dong Zhang and Chi-Yin Chow.
\newblock Geosoca: Exploiting geographical, social and categorical correlations
  for point-of-interest recommendations.
\newblock In {\em Proceedings of the 38th international ACM SIGIR conference on
  research and development in information retrieval}, pages 443--452, 2015.

\bibitem{zhang2018GraphInception}
Yizhou Zhang, Yun Xiong, Xiangnan Kong, Shanshan Li, Jinhong Mi, and Yangyong
  Zhu.
\newblock Deep collective classification in heterogeneous information networks.
\newblock In {\em Proceedings of the 2018 World Wide Web Conference}, pages
  399--408, 2018.

\bibitem{zitnik2018modeling}
Marinka Zitnik, Monica Agrawal, and Jure Leskovec.
\newblock Modeling polypharmacy side effects with graph convolutional networks.
\newblock {\em Bioinformatics}, 34(13):i457--i466, 2018.

\end{thebibliography}

\newpage
\appendix
\textbf{\Large Supplementary Material}

This supporting material contains additional dataset description, GripNet implementations \& variants, baseline implementations, additional performance comparison, data integration results and ablation studies. 

\section{Further information on datasets}
\label{app:dataset}

\subsection{PoSE-series}
As shown in Table~\ref{tab:pose}, we construct three large-scale datasets for polypharmacy side effect prediction task defined in \cite{zitnik2018modeling} by integrating three BioSNAP-Decagon sub-datasets provided by \cite{snap}: 
\begin{itemize}
	\item \textbf{PP (gene-gene)}: relations between genes depending on whether the proteins expressed by them have functional association or physical connection.
	\item \textbf{GhG (drug-gene)}: relations between drugs and their targeted proteins (genes).
	\item \textbf{ChChSe (drug-drug)}: interactions between drugs labelled with polypharmacy side effects. 
\end{itemize}

The difference between PoSE-0, 1 and 2 lies in the number of gene nodes that are not linked with drug nodes. We evaluate GripNet on them to test its performance with different proportions of indirect task-related information.
The PoSE-2 integrates all the data from the above datasets, while the PoSE-0 ( PoSE-1) removes gene (drug) nodes that are not  directly connected by drug (gene) nodes.
\begin{itemize}
	\item \textbf{PoSE-0} All genes are linked with drugs, while 361 drugs are not linked with any gene
	\item \textbf{PoSE-1} 15,441 genes are not linked with any drug, while all drugs are linked with genes.
	\item \textbf{PoSE-2} 15,441 genes and 361 drugs are not linked with any other kind of entity.
\end{itemize}

\begin{table}[h]
	\caption{PoSE Dataset Statistics. \textbf{$|{V}_{g/d}|$}: number of gene/drug nodes, \textbf{$|{E}_{gg/gd/dd}|$}: number of gene-gene/gene-drug/drug-drug edges, $|\mathcal{L}|$: number of drug-drug edge labels (i.e. polypharmacy side effects), \textbf{AN}: average number of drug-drug edges with a common side effect label. \textbf{NG/ND}: number of gene/drug nodes that are not directly linked with any drug/gene nodes.}
	\label{tab:pose}
	\centering
	\scalebox{1}{
		\begin{tabular}{lccccccccc}
			\toprule
			Dataset     & \textbf{$|{V}_g|$}     & \textbf{$|{V}_d|$} & \textbf{$|{E}_{gg}|$}     & \textbf{$|{E}_{gd}|$} & \textbf{$|{E}_{dd}|$}     & $|\mathcal{L}|$ & \textbf{AN} & \textbf{NG}  & \textbf{ND}\\
			\midrule
			PoSE-0 & 3,640 & 645 & 81,486 & 18,596 & 4,625,608 & 1,097 & 4,217 & 0 &  361 \\
			PoSE-1 & 19,081 & 284 & 1,431,224 & 18,596 & 1,171,603 & 861 & 1,361 & 15,441 & 0 \\
			PoSE-2 & 19,081 & 645 & 1,431,224 & 18,596 & 4,625,608 & 1,097 & 4,217 & 15,441 & 361 \\
			
			\bottomrule
	\end{tabular}}
\end{table}

\subsection{Aminer}
Aminer is constructed from the  Aminer Academic Network dataset in \cite{Tang:08KDD}. Following the top venue classification in Google Scholar (\url{https://scholar.google.com/citations?view_op=top_venues&hl=en&vq=eng_enggeneral}), we select seven categories in computer science as node attributes to be predicted: 1) Computer Vision \& Pattern Recognition, 2) Data Mining \& Analysis, 3) Computer Security \& Cryptography, 4) Computational Linguistics, 5) Computer Networks \& Wireless Communication, 6) Computer Graphics, and 7) Theoretical Computer Science.

\subsection{Freebase-Series}

\begin{table}[h]
	\caption{Freebase Dataset Statistics. \textbf{$|{E}_{bo\_bo}|$}: number of book-book edges, \textbf{$|{E}_{bo\_l}|$}: number of book-location edges, \textbf{$|{E}_{bo\_o}|$}: number of book-organization edges, \textbf{$|{E}_{ll}|$}: number of location-location edges, \textbf{$|{E}_{oo}|$}: number of organization-organization edges, \textbf{$|{E}_{lo}|$}: number of location-organization edges, \textbf{$|{E}_{bo\_bu}|$}: number of book-business edges, \textbf{$|{E}_{bu\_bu}|$}: number of business-business edges.}
	\label{tab:freebase_ap}
	\centering
	\scalebox{1}{
		\begin{tabular}{lccccccccc}
			\toprule
			Dataset     & \textbf{$|{E}_{bo\_bo}|$}     & \textbf{$|{E}_{bo\_l}|$} & \textbf{$|{E}_{bo\_o}|$}     & \textbf{$|{E}_{ll}|$} & \textbf{$|{E}_{oo}|$}  & \textbf{$|{E}_{lo}|$} & \textbf{$|{E}_{bo\_bu}|$}  & \textbf{$|{E}_{bu\_bu}|$}\\
			\midrule
			Freebase-a & 12,556 & - & - & - & - & - & - & - \\
			Freebase-b & 12,556 & - & - & - & - & - & 24,644 & 810,832 \\
			Freebase-c & 12,556 & - & 21,507 & - & 129,124 & - & 24,644 & 810,832 \\
			Freebase-d & 12,556 & 58,466 & 21,507 & 2,138,818 & 129,124 & 158,132 & 24,644 & 810,832 \\
			
			\bottomrule
	\end{tabular}}
\end{table}

Table \ref{tab:freebase_ap} shows the detailed Freebase series dataset statistics. We extract a subset containing 14,989 books, 339,972 businesses, 102,543 organizations and 642,896 locations. Each book node is labelled with one of the eight genres of literature, 
and has at least one edge linked with one node of another type. In order to explore GripNet's performance on datasets of different scale, we construct four networks from the extracted Freebase data:
\begin{itemize}
	\item \textbf{Freebase-a} is a homogeneous graph with only book nodes and links between them. 
	\item \textbf{Freebase-b} is a heterogeneous graph containing book and business nodes and three types of edges (book-book, book-business and business-business). 
	\item \textbf{Freebase-c} contains an additional node type (organization) on the basis of Freebase-b, and  three additional types of edges (book-book, book-business and business-business). 
	\item \textbf{Freebase-d} involves all four types of nodes and eight types of edges.
\end{itemize}

\section{Additional notes on Implementation}

\subsection{GripNet Implementation}
In this section, we explain the GripNet models with different number of supervertices together. The input features of the entities mentioned below are all one-hot encoding.

\begin{wrapfigure}{r}{0.2\textwidth}
	\vspace{-8mm}
	\vspace{2mm}
	\hfill
	\includegraphics[width=0.2\columnwidth]{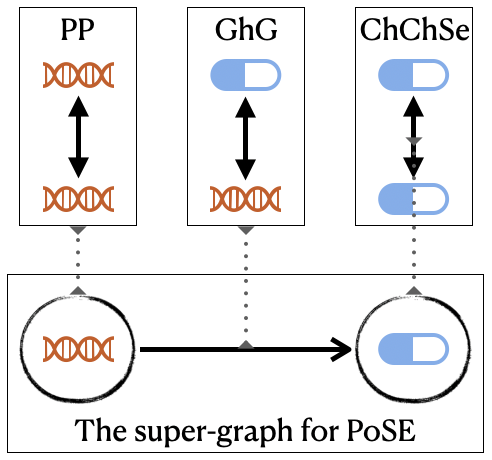}
	\caption{Supergraph for PoSE-series datasets}
	\label{fig:posegraph}
	\vspace{-6mm}
\end{wrapfigure}
\vspace{-2mm}
\subsubsection{GripNet-0, -1, and -2} 
GripNet-0, -1, and -2 are GripNet implementations for polypharmacy side-effect prediction task on datasets PoSE-0, -1 and -2. 
Usually the prediction is made directly on the drug-drug interaction graph in the ChChSe dataset. Assuming that drug-drug interactions are caused by interactions between proteins (genes) affected by them, integrating the PP and GhG graph may improve the prediction performance. In Figure \ref{fig:posegraph}, we show the supergraph for PoSE datasets and define the information propagation path for our model. Our model therefore 1) learns the protein embeddings in the protein supervertex (P) with $\mathbf{W}^{(in)} \in \mathbb{R}^{|P|\times 32}, \mathbf{W}_1^{(ia)} \in \mathbb{R}^{32\times 16}$, and $ \mathbf{W}_2^{(ia)} \in \mathbb{R}^{16\times 16}$, and 2) propagates the information to the drug supervertex (D) via an external aggregation feature layer with $\mathbf{W}^{(ex)} \in \mathbb{R}^{16\times 16}$ and learns the drug embeddings for the side effect prediction on the drug supervertex with $\mathbf{W}^{(in)} \in \mathbb{R}^{|D| \times 32}$, and $ \mathbf{W}^{(ia)} \in \mathbb{R}^{(32+16) \times 16}$.

\subsubsection{GripNet-b, -am}  They are similar to GripNet-0/1/2 as their supergraph also contains only 2 supervertices with one directed edge. For GripNet-b on Freebase-b dataset, the business node embeddings are learned on the business supervertex (Bu) with $\mathbf{W}^{(in)} \in \mathbb{R}^{|Bu|\times 128}, \mathbf{W}_1^{(ia)} \in \mathbb{R}^{128. \times 64}$, and $ \mathbf{W}_2^{(ia)} \in \mathbb{R}^{64\times 64}$, then they are propagated to the book supervertex (Bo) with $\mathbf{W}^{(ex)} \in \mathbb{R}^{64\times 64}$. The final book embeddings are learned with $\mathbf{W}^{(in)} \in \mathbb{R}^{|Bo| \times 128}, \mathbf{W}^{(ia)}_1 \in \mathbb{R}^{(128+64) \times 64}$, and $ \mathbf{W}^{(ia)}_2 \in \mathbb{R}^{64 \times 32}$. 

For GripNet-am on Aminer dataset, the paper node embeddings are learned on the paper supervertex (P), then they are propagated to the author  supervertex (A). The final author embeddings are learned within A. GripNet-am have the same trainable parameter settings and dimensions except those of the internal aggregation layer on the target supervertex, which has two sub-layers with $\mathbf{W}^{(ia)}_1 \in \mathbb{R}^{(128+64) \times 128}$, and $\mathbf{W}^{(ia)}_2 \in \mathbb{R}^{128 \times 32}$.

\subsubsection{GripNet-c, -d}
The Figure 1 in the main text gives an examples on model construction for book classification on Freebase-d dataset from a heterogeneous graph segregation perspective. For Freebase-d, there are \textbf{three supervertices}: book $v_0^S$, location\&organization $v_1^S$, and business $v_2^S$. The \textbf{information propagation paths} are: location\& organization $\rightarrow$ book, and business $\rightarrow$ book. In each supervertex, the input node features of all the entities use the one-hot encoding, and the output is the node embeddings. The location, organization and business node embeddings are learned with 1) a internal feature layer with  $\mathbf{W}^{(in)} \in \mathbb{R}^{|V_i|\times 256}$,  where $V_i$ are nodes in the $i^{th}$ supervertex, and 2) two internal aggregation layers with $\mathbf{W}^{(ia)}_1 \in \mathbb{R}^{256 \times 128}$, and $\mathbf{W}^{(ia)}_2 \in \mathbb{R}^{128 \times 128}$. On the book supervertex, 1) the business, location and organization embeddings are transform into book features by an external aggregation feature layer with $\mathbf{W}^{(ex)}_{1}, \mathbf{W}^{(ex)}_{2} \in \mathbb{R}^{128 \times 128}$, 2) the input feature is extracted in a internal feature layer with  $\mathbf{W}^{(in)} \in \mathbb{R}^{|V_0|\times 128}$, and 3) the final book embeddings are leaned by an internal aggregation layer with $\mathbf{W}^{(ia)}_1 \in \mathbb{R}^{256 \times 32}$, whose input is the concatenation of the output of external and internal feature layers.

The supergraph construction for Freebase-c is similar to that for Freebase-d. It also contains \textbf{three supervertices}: book $v_0^S$, organization $v_1^S$, and business $v_2^S$. The \textbf{information propagation paths} are: organization $\rightarrow$ book, and business $\rightarrow$ book. GripNet implementation for Freebase-c is the same as that for Freebase-d, as they have the same layer settings and dimensions of all the trainable parameters.

\subsubsection{GripNet-a}
The Freebase-a is a special dataset, as it is represented as a homogeneous graph. It is only used for data integration. The supergraph of GripNet-a only contains a supervertex - the book supervertex, and no superedges. On the book (Bo) supervertex, the embeddings are learned without external aggregation feature layer. The parameter settings are $\mathbf{W}^{(in)} \in \mathbb{R}^{|Bo| \times 128}, \mathbf{W}^{(ia)}_1 \in \mathbb{R}^{(128) \times 64}$, and $ \mathbf{W}^{(ia)}_2 \in \mathbb{R}^{64 \times 32}$.

\subsection{Baselines Implementation}
The implementation of \textbf{GCN} in \cite{kipf2017gcn}, \textbf{RGCN} in \cite{schlichtkrull2018modeling}, \textbf{DistMult} in \cite{yang2014embedding} and \textbf{GAT} in \cite{velickovic2018graph} models on node classification are provided by PyTorch-Geometric in \cite{pytorch}. For the GAT model, we pick the number of the self-attention head to be 4, and the number of features to be 16. For the GCN model, we use 2 convolutional layers with hidden dimension 64. For the RGCN model, the number of layers is 2 and the input, hidden and output dimensions are all set to be 32.

\section{Additional Performance Comparison}

We compare the model performance on prediction accuracy, peak GPU memory usage and time cost with five baselines: \textbf{TransE} in \cite{bordes2013transE}, \textbf{RotatE} in \cite{sun2019rotate}, \textbf{ComplEx} in \cite{trouillon2016complex}, \textbf{DistMult} \cite{yang2014embedding} and \textbf{RGCN} \cite{schlichtkrull2018modeling}, when leaning embeddings with same dimensions on the PoSE-0 and PoSE-2 datasets. We have already discussed the model accuracy comparison based on Figure 2B in the main paper which is plotted according to the following Table \ref{tab:dim2}. 

\begin{table}[h]
	\caption{Multi-relational Link prediction results in \textbf{AUROC} on the PoSE-2 dataset. The best result is in bold and second best one is underlined. Analogous trends hold for AUPRC and AP@50. \textbf{Dim-n}: learning the fix n-dimensional embeddings for each model. \textbf{Mem}: peak GPU memory usage; \textbf{TpE}: computational time per epoch (including training and testing score computation).}
	\label{tab:dim2}
	\centering
	\scalebox{0.85}{
		\begin{tabular}{lrrrrrrrrrrrr}
			\toprule
			& \multicolumn{3}{c}{Dim-8} & \multicolumn{3}{c}{Dim-16}  & \multicolumn{3}{c}{Dim-24} & \multicolumn{3}{c}{Dim-32}  \\
			\cmidrule(r){2-4} \cmidrule(r){5-7} \cmidrule(r){8-10} \cmidrule(r){11-13}
			Model     & \textbf{AUROC}     & \textbf{Mem} & \textbf{TpE(s)}     & \textbf{AUROC} & \textbf{Mem}     & \textbf{TpE(s)} & \textbf{AUROC} & \textbf{Mem}     & \textbf{TpE(s)} & \textbf{AUROC} & \textbf{Mem}     & \textbf{TpE(s)}\\
			\midrule
			TransE   & 0.591 & 2.81 & 29.1 & 0.652 & 3.95 & 29.3 & 0.669 & 5.41 & 29.8 & 0.674 & 9.64 & 29.9\\
			
			RotatE   & 0.696 & 10.03 & 30.1 & 0.796 & 19.43 & 29.7 & 0.851 & 29.98 & 30.6 & - & - & -\\
			ComplEx  & 0.501 & 9.33 & 29.4 & 0.488 & 18.12 & 29.3 & 0.469 & 26.89 & 28.7 & - & - & -\\
			DistMult & 0.498 & 4.16 & 29.7 & 0.503 & 7.74 & 29.2 & 0.517 & 11.31 & 29.2
			& 0.800 & 14.89 & 29.6\\
			RGCN     & 0.888 & 19.92 & 54.5 & 0.880 & 23.35 & 54.9 & 0.888 & 26.93 & 55.1 & 0.846 & 26.63 & 55.3\\
			\midrule
			\textbf{GripNet-l}  & 0.921 & 21.18 & 40.14 & 0.918 & 23.91 & 40.3 & 0.917 & 26.64 & 40.8 & 0.918 & 29.03 & 41.2\\
			\bottomrule
	\end{tabular}}
\end{table}

As shown in Table \ref{tab:dim2} and Figure 2A in the main paper, when leaning with a fixed embedding dimension, we have the following observations:
\begin{itemize}
	\item Sensitivity: \textbf{GripNet} < TransE < RGCN < RotatE, ComplEx, and DistMult 
	\item Training time per epoch: TransE, RotatE, ComplEx and DistMult < \textbf{GripNet} < RGCN
	\item Overall converge speed: TransE > \textbf{GripNet} > DistMult > RotatE > ComplEx and RGCN
	\item Memory cost: TransE < DistMult < RGCN < \textbf{GripNet} < ComplEx and RotatE
\end{itemize}

\section{Further Data Integration Result}
\label{app:inte}
\subsection{Polypharmacy side effect prediction}

For the polypharmacy side effect prediction task which predicts the side effects of drug pairs, Table \ref{tab:int-pose} shows that 1) Pharmacological information does contain drug-drug interaction information, as we can get decent result by using it directly in the PP \& GhG dataset, and 2) additional pharmacological information in the model for the PoSE-2 dataset only improves the performance of the one for the ChChSe dataset slightly with 1.5\%. 
\begin{table}[h]
	\caption{Results of GripNet for Polypharmacy side effect prediction. \textbf{g-g}, \textbf{g-d} and \textbf{d-d} are interaction information between genes, gene-drug and drugs respectively.}
	\label{tab:int-pose}
	\centering
	\scalebox{0.88}{
		\begin{tabular}{ccc}\\\toprule  
			\textbf{Dataset} & \textbf{Information} & \textbf{AUROC} \\\midrule
			PP \& GhG  & g-g \& g-d & 0.743\\  \midrule
			ChChSe &d-d & 0.908\\  \midrule
			PoSE-2 & g-p, g-d \& d-d & 0.922\\  \bottomrule
	\end{tabular}}
\end{table}

\subsection{Book label classification}
For this task, as shown in Table \ref{tab:int-book}, the classification accuracies of GripNets on Freebase-b, -c, and -d (where additional information is added) are significantly better than that of GripNet on Freebase-a (where only book information is included), by 58.7\%, 87.7\&, and 99.2\%.

\begin{table}[h]
	\caption{Results of GripNet for book classification. \textbf{b}, \textbf{bus}, \textbf{loc} and \textbf{org} are information related book, business, location and organization respectively.}
	\label{tab:int-book}
	\centering
	\scalebox{0.88}{
		\begin{tabular}{ccc}\\\toprule  
			\textbf{Dataset} & \textbf{Information} & \textbf{Micro-F1} \\\midrule
			Freebase-a  & b & 0.300\\  \midrule
			Freebase-b & b \& bus & 0.476\\  \midrule
			Freebase-b & b \& bus \& org & 0.563\\  \midrule
			Freebase-d & b, bus, loc \& org & 0.597\\  \bottomrule
	\end{tabular}}
\end{table}

\section{Ablation Study}
\label{app:ab}

We perform an ablation study to understand the influence of different hyper-parameter and design detail choices to our results.

\subsection{Input to the internal aggregation layer}
As shown in the first two rows of Table \ref{tab:an}, taking $[\mathbf{r}^c_{ex},\mathbf{r}^c_{in}]$ or $\mathbf{r}^c_{ex} + \mathbf{r}^c_{in}$ as the input of the internal aggregation layer have similar performance. The construction of GripNet models is flexible, as the vector dimensions of representing information propagated from other supervertices and from node attributes within this supervertex can be different. When the current supervertex wants to make the dimensions of the information sent from other nodes more or less than those from its attribute, the concatenation operation (default operation) is a better choice. For GripNet-add on Freebase-b dataset, the dimensions of the information from its attribute are double of that from other supervertex, while the dimensions of them are the same in GripNet-cat. Results show that controlling the ratio of node input attribute to the incoming information can change the model performance.

\begin{table}[h]
	\caption{Node classification results in \textbf{Micro-F1}. The best result is in bold and second best one is underlined. Analogous trends hold for Macro-F1. \textbf{Mem}: peak GPU memory usage; \textbf{TpE}: computational time per epoch (including training and testing score computation).}
	\centering
	\scalebox{0.88}{
		\centering
		\begin{tabular}{lrrrrrrrrrrrr}
			\toprule
			& \multicolumn{3}{c}{Aminer} & \multicolumn{3}{c}{Freebase-b}  & \multicolumn{3}{c}{Freebase-c} & \multicolumn{3}{c}{Freebase-d} \\
			\cmidrule(r){2-4} \cmidrule(r){5-7} \cmidrule(r){8-10} \cmidrule(r){11-13}
			Model     & \textbf{Mi-F1}     & \textbf{Mem} & \textbf{TpE(s)}    & \textbf{Mi-F1} & \textbf{Mem} & \textbf{TpE(s)}      & \textbf{Mi-F1} & \textbf{Mem} & \textbf{TpE(s)}  & \textbf{Mi-F1} & \textbf{Mem} & \textbf{TpE(s)}   \\
			\midrule
			
			\textbf{GripNet} & 0.920 & 5.89 & \textbf{0.25}& 0.464 & \textbf{2.29} & \textbf{0.08} & 0.563 & \textbf{4.45}& \textbf{0.16}& 0.592& 14.30 &\textbf{0.36}\\
			
			GripNet-add & 0.921 & 5.74 &\textbf{0.25} & \textbf{0.476} & 4.63 &0.13 & \textbf{0.557}& 5.18& 0.17& 0.597& 14.26 &\textbf{0.36}\\
			\midrule
			
			GripNet-cat-lin & 0.920 & 5.89 &\textbf{0.25} & 0.466 & \textbf{2.29} &\textbf{0.08} & 0.563  & 4.48 &0.17 & \textbf{0.599}& \textbf{14.08} &\textbf{0.36}\\
			GripNet-add-lin & \textbf{0.922} & \textbf{5.71} &0.26 & 0.462 & 4.61 & 0.13  & 0.551& 5.12 &0.17 & 0.584 & 14.26 &\textbf{0.36}\\
			\bottomrule
	\end{tabular}}
	\label{tab:an}
\end{table}

\subsection{Transformation between supervertices}
We explore the GripNet-cat and GripNet-add models with and without a non-linear transformation ReLU in Equ.(2) on the Aminer, Freebase-b, -c, and -d datasets. Results show in the last two rows in Table \ref{tab:an}, using nonlinear transformation between different super nodes is not always better than using a linear transformation.

\subsection{Categorized negative sampling (CNS)} GripNet only needs CNS when modelling relations within the drug supervertex. When we replaced it with the popular negative sampling used in comparative methods for link prediction (RGCN, DistMult and DECAGON), the accuracy on the PoSE-series datasets was decreased by no more than 1\% (see Table \ref{tab:mrlp}), but the convergence speed slowed down a little.

\begin{table}[h]
	\caption{Multi-relational Link prediction results in \textbf{AUROC}. The best result is in bold and second best one is underlined. Analogous trends hold for AUPRC and AP@50. \textbf{Mem}: peak GPU memory usage; \textbf{TpE}: computational time per epoch (including training and testing score computation).}
	\label{tab:mrlp}
	\centering
	\scalebox{0.87}{
		\begin{tabular}{lrrrrrrrrr}
			\toprule
			& \multicolumn{3}{c}{PoSE-0} & \multicolumn{3}{c}{PoSE-1}  & \multicolumn{3}{c}{PoSE-2}  \\
			\cmidrule(r){2-4} \cmidrule(r){5-7} \cmidrule(r){8-10}
			Model     & \textbf{AUROC}     & \textbf{Mem} & \textbf{TpE(s)}     & \textbf{AUROC} & \textbf{Mem}     & \textbf{TpE(s)} & \textbf{AUROC} & \textbf{Mem}     & \textbf{TpE(s)}\\
			\midrule
			
			\textbf{GripNet}-noCNS  & 0.920 & 15.58 & 35.0 & 0.911 & 12.18 & 19.3 & 0.919 & 15.84 & 35.0 \\
			\textbf{GripNet}  & \textbf{0.922} & 15.61 & 35.0 & \textbf{0.915} & 12.22 & 19.3 & \textbf{0.922} & 15.96 & 35.1 \\
			\bottomrule
	\end{tabular}}
\end{table}

\end{document}